# Bangla BERT for Hyperpartisan News Detection: A Semi-Supervised and Explainable AI Approach

Mohammad Mehadi Hasan*, Fatema Binte Hassan†, Md Al Jubair†,
Zobayer Ahmed‡, Sazzatul Yeakin§, Md Masum Billah¶
*Department of CSE, Bangladesh University of Business and Technology, Dhaka, Bangladesh
hasanbdmehadicse@gmail.com
†Department of CSE, Chittagong University of Engineering and Technology, Chittagong, Bangladesh
fatemabintehassan@gmail.com, aljubairpollob@gmail.com
‡Department of CSE, European University of Bangladesh, Dhaka, Bangladesh
zobayer115@gmail.com
§Founder, Cosmic IT, Dhaka, Bangladesh
cosmic.syc.3@gmail.com
¶Founder, Kidz Programming, Dhaka, Bangladesh
masum.swe.ndc@gmail.com

*Abstract*—In the current digital landscape, misinformation circulates rapidly, shaping public perception and causing societal divisions. It is difficult to identify hyperpartisan news in Bangla since there aren't many sophisticated natural language processing methods available for this low-resource language. Without effective detection methods, biased content can spread unchecked, posing serious risks to informed discourse. To address this gap, our research fine-tunes Bangla BERT. This is a state-of-the-art transformer-based model, designed to enhance classification accuracy for hyperpartisan news. We evaluate its performance against traditional machine learning models and implement semi-supervised learning to enhance predictions further. Not only that, we use LIME to provide transparent explanations of the model's decision-making process, which helps to build trust in its outcomes. With a remarkable accuracy score of 95.65%, Bangla BERT outperforms conventional approaches, according to our trial data. The findings of this study demonstrate the usefulness of transformer models even in environments with limited resources, which opens the door to further improvements in this area.

## I. Introduction

Online news consumption has surged exponentially as digital media's explosive expansion unfolds. But this has also resulted in the proliferation of hyperpartisan material, which frequently advances biassed or false information. Maintaining the integrity of information ecosystems and guaranteeing that readers have access to fair and accurate news depend on the detection of such material. Although hyperpartisan content in high-resource languages like English [1] has been much improved upon, low-resource languages like Bangla still get little research attention. Developing good detection models in these languages is much challenged by the absence of labelled datasets and language-specific tools.

Prior research mostly concentrated on conventional machine learning methods such as Random Forest and Logistic Regression for hyperpartisan news identification. Although these models show potential, they frequently find it difficult to adequately recognise hyperpartisan material by capturing the contextual subtleties of language. Because they can grasp difficult semantic linkages, transformer-based models such as BERT [2] have shown outstanding performance in high-resource languages. But little study has been done on its applicability to low-resource languages like Bangla, therefore creating a gap in the knowledge. Moreover, the absence of explainability in these models reduces their transparency and dependability, which is very vital for practical application.

In this work, we propose a fine-tunized Bangla BERT model to solve the hyperpartisan news identification difficulty in Bangla. We evaluate its performance among traditional machine learning models like Support Vector Machine, Random Forest, and Logistic Regression. We apply Explainable AI (XAI) methods [3], including LIME (Local Interpretable Model-agnostic Explanations), to offer insights on the model's decision-making process, thereby improving transparency and interpretability. Our results show Bangla BERT offers interpretable predictions in addition to state-of- the-art performance in accuracy, precision, recall, and F1-score surpassing conventional models.

To the best of our knowledge, this is the first study to detect hyperpartisan news content in Bangla using a semi-supervised fine-tuning of Bangla BERT. Furthermore, we are the first to apply LIME explainability in this context to provide interpretable outputs for Bangla text classification. This work thus establishes a strong initial benchmark and opens a new research direction in low-resource NLP.

This paper is structured as follows generally: Section II addresses the related work; Section III explains the technique; Section IV shows the experimental results; Section V finishes the study with future directions. This study guarantees openness and trustworthiness by using Bangla BERT's strengths, conventional machine learning models, and XAI approaches to add to the expanding corpus of research on hyperpartisan news identification in low-resource languages.

## II. Related Work

Recently, It becomes a common scenario that biased information is affecting the society a lot. Researchers are finding numerous ways, such as transformer-based models, traditional machine learning models, and neural network models. Using progress in natural language processing (NLP), machine learning, and social network analysis, these efforts aim to automatically find content that is biassed.

Transformer-based models have become the most popular way to find hyperpartisan news because they can pick up on details in text that come from context. An automatic model was created using BERT, ELMo, and Word2Vec embeddings [4]. ELMo paired with a Random Forest classifier gave the best results (88% accuracy). Their work showed that BERT could figure out what local words meant without needing to be fed the whole piece, which saved a lot of time and effort. Also, [5] used masking methods with transformer models such as BERT, M-BERT, and XLM-RoBERTa. They discovered that topic-related features worked better than style-based ones. However, the problem is, it takes too much time. The most useful part for classifying was found to be the beginning of news stories. SemEval-2019 Task 4 [1] confirmed that transformer-based models work even better, with CNN + ELMo getting the best results (82.2%). But the competition showed how flawed publisher-labeled datasets are because they add noise because publishers cross in training and testing splits.

Transformers, feature engineering, and traditional machine learning algorithms may detect hyperpartisan news. Agerri [6] used clustering-based semantic features and shallow local features for semi-supervised learning and achieved 76.1% accuracy. Using dependency sub-trees and named entity recognition (NER) tags, Nguyen et al. [7] achieved 78.7% accuracy. Their research showed that grammatical characteristics differentiate hyperpartisan text. A linear classifier using feature engineering by Hanawa et al. [8] used BERT embeddings, article length features, and informative phrases. Their ensemble model placed third in SemEval-2019 with 80.9% accuracy. Although feature-based techniques were successful, simpler models like bag-of-words (BoW) with logistic regression performed badly on hyperpartisan categorisation (65.6% accuracy) but well on publisher-based tasks (70.6% accuracy) [9]. Overfitting from publisher-specific traits may impede generalisation to unseen data.

Social media is a great environment for researching hyperpartisanship and its effects. Recuero et al. [10] examined over 8 million tweets from the 2018 Brazilian presidential election and found hyperpartisan news sites dominated the debate. During the runoff, misinformation grew in pro- and anti-Bolsonaro clusters, according to their findings. This study showed how social media fuels political polarisation and disinformation. Lyu et al. [11] computationally analysed 1.8 million news headlines from 2014 to 2022 and found that right-leaning media utilised more hyperpartisan names. During the 2016 U.S. presidential election, left-leaning media saw a rise in hyperpartisan titles.

These data show hyperpartisanship's dynamic character and association with key political events. [12] used a German dataset for fine-grained political bias categorisation and found that publisher overfitting made binary models (79% accuracy) outperform five-class models (43% accuracy). They stressed the need of article-level annotations for model robustness.

Despite these advances, hyperpartisan news identification remains difficult. Dataset bias is a serious issue in publisher-labeled corpora like SemEval-2019 [1]. Noise and overlapping publications in training and testing divides make such dataset models difficult to generalise [1], [13]. Transformer-based models like BERT [4], [8] demand large computational resources for training and inference, restricting their real-time implementation. Most studies ignore multilingual and regional bias dynamics by focussing on English [1], [4], [5], [10], [11], [14]–[16] or German [12]. This restricts model applicability to varied cultural and language environments. Feature interpretability also needs consideration. Attention processes [17] and Shapley values [11] increase model explainability, although bias words [16] are understudied. Understanding these patterns might improve model performance and reveal hyperpartisanship's language indicators.

Innovative methods have been used to close these gaps. [13] reduced noise in poorly labelled datasets with semi-supervised pseudo-labeling and a fine-tuned BERT model to 75.8% accuracy. Their findings showed that huge, noisy datasets can enhance categorisation. Joo and Hwang [18] suggested a 74.5%-accurate ensemble learning model using gradient boosting decision trees (GBDT) and N-gram CNNs. Their approach showed that style-based and content-based characteristics improve hyperpartisan identification. Huang and Lee [16] added bias word frequency, which improved classification accuracy by 6% using ELMo embeddings. These advances demonstrate how hybrid models and feature engineering overcome dataset restrictions and improve model performance.

Finally, all the ai related approaches have improved the detection of hyperpartisan news. Language barriers, computational costs, and biased datasets are still a major problems. Robustness, language free models should be the topic of the future research. These problems might be solved with the use of ensemble models and semi-supervised learning. By using these method we can fight against divided society.

## III. Data Preparation

We follow four key steps: dataset collection, data pre-processing, data augmentation, and data description prior to feeding the data into the model.

### A. Data Collection

In today's digital age, data is widely available online. However, compared to many other languages, Bangla remains a low-resource language, making data collection particularly challenging. To address this issue, the authors utilised the BNAD dataset [19]), which includes news articles from eight Bangla news websites, such as ManabZamin and Kalerkantha. The dataset contains over 1.9 million articles across various topics, including politics, sports, economy, technology, and more. For this study, the authors specifically extracted political news articles, resulting in a subset of 48,831 entries, which served as the primary data set for their research.

Table I
SAMPLE EXAMPLES FROM DATASET

| Title | Content | Source |
|---|---|---|
| বর্ণচোরা মুক্তিযোদ্ধাদের প্রতিহত করাই আজকের শপথ | স্বাধীনতার শত্রু, সাম্প্রদায়িক অপশক্তি এবং বর্ণচোরা মুক্তিযোদ্ধাদের প্রতিহত করে তাদেরকে চূড়ান্তভাবে পরাজিত করাই হচ্ছে আজকের | AjkerPatrika |
| হজে হতাহতের ঘটনা তদন্তের দাবি দেশের ধর্মভিত্তিক রাজনৈতিক দলগুলোর | সৌদি আরবে হজের সময় ক্রেন দুর্ঘটনা এবং মিনায় পদদলিত হয়ে নিহত হওয়ার ঘটনার তদন্ত দাবি করেছে | BanglaTribune |

### B. Data Preprocessing

After gathering the data, the first step involves ensuring data quality by dropping rows with empty 'Title' or 'Content' fields. This is crucial to address missing or incomplete data, as empty rows can negatively impact the analysis. Next, the 'Title' and 'Content' columns are combined into a single 'text' column to create a unified input for further processing. This consolidation simplifies the handling of textual data.

Once the data is consolidated, the text is cleaned by removing special characters, numbers, and symbols to improve the quality of the dataset. This step ensures that the text is standardized and ready for deeper analysis. Following this, the cleaned text is tokenized, breaking it down into smaller components, or tokens, which represent individual words or phrases.

After tokenization, Bangla stopwords—commonly used words that add little meaning to the text—are removed to focus on the more meaningful tokens. Finally, the text is then stemmed, a process that reduces words to their root or base form, further simplifying the dataset. This entire process ensures that the data is clean, structured, and ready for effective analysis or modeling.

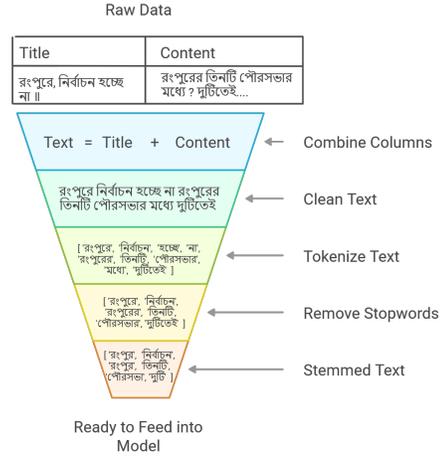

Figure 1. Data Preprocessing

### C. Data Augmentation

The initial labeling of 805 articles was insufficient. To expand the dataset, we used data augmentation by translating Bangla text to English and back to Bangla three times using the GoogleTrans package. This process increased the data set to 3,220 samples.

### D. Dataset Description

The dataset is composed of two parts: a labeled set with 3,200 samples (1,476 hyperpartisan and 1,744 not hyperpartisan) and an unlabeled set with 47,987 samples, both obtained after preprocessing and data augmentation. Although the labeled data is limited in size, we adopt a semi-supervised learning approach to overcome this challenge. By applying pseudo-labeling and confidence-based filtering to the unlabeled data, we enhance the model's ability to detect hyperpartisan content.

## IV. Methodology

The researchers aimed to detect hyperpartisan Bangla news articles using a semi-supervised learning approach. Initially, they trained the model on a small labeled dataset. Once the model was trained, it was used to label an unlabeled dataset, which was then incorporated into the training process. So, the model must perform very well during the training using labeled dataset.

This study framed the problem as a classification task, where the goal was to predict two distinct classes from Bangla news articles. Selecting an optimal model was challenging, as it needed to comprehend the overall context of each article. To address this, the researchers experimented with two different approaches: traditional machine learning methods and a transformer-based model, Bangla BERT. In both cases, hyperparameter tuning was essential to optimize performance.

For the traditional machine learning approach, the researchers experimented with algorithms such as Logistic Regression, Random Forest Classifier, Naïve Bayes Classifier, and Support Vector Machine. The text data was first vectorized using the Term Frequency-Inverse Document Frequency (TF-IDF) method. The vectorizer was fitted on the training data and then used to transform both the training and test datasets. The dataset was split into training (70%), validation (15%), and test (15%) sets. After preprocessing, the transformed data was fed into the traditional machine learning models. To optimize these models, the researchers used GridSearchCV to determine the best hyperparameter values. The best parameter values, are presented in Table 2.

Table II
BEST PARAMETERS AFTER GRIDSEARCHCV

| Logistic Regression | Random Forest Classifier | Naive Bayes Classifiers | Support Vector Machine |
|---|---|---|---|
| C : 10 | bootstrap : False | alpha : 0.01 | C : 10 |
| max_iter: 100 | max_depth : None | fit_prior : True | $coef0$ : 0.1 |
| penalty : $l2$ | min_samples_leaf : 1 | - | degree : 3 |
| solver : saga | min_samples_split : 10 | - | gamma : scale |
| - | n_estimators : 200 | - | kernel : poly |

For any BERT model, there has an intrinsic limitation of a maximum of 512 word tokens. Only the first 500 word token of the articles will be used as the sequence length to include more context and details from the articles. The pre-trained Bangla BERT model was trained using 70% of the training data. The model took a sequence of words as input and passed them through a transformer neural network. This process created word embeddings, which are special number representations for each word based on the context of the text. These embeddings were then used to train the Bangla BERT classifier in the final layer to predict the class of each article. After training, the model was tested. Finally, the trained Bangla BERT model was used to label the unlabeled dataset and train again. The researchers measured how well the model performed using precision, recall, and F1-score. It was the best model till now. It gives 95.65% accuracy.

## V. EXPERIMENT AND RESULT ANALYSIS

### A. Performance Evaluation

The performance evaluation of the implemented models for hyperpartisan news detection in Bangla text reveals that Bangla BERT, a transformer-based model, outperforms traditional machine learning models across all metrics. Bangla BERT achieved the highest accuracy (95.65%), precision (95.65%), recall (95.24%), and F1-score (95.44%), demonstrating its ability to capture contextual nuances and semantic relationships in text. Among the traditional models, Random Forest Classifier and Support Vector Machine (SVM) emerged as the top performers, with F1-scores of 93.55% and 93.27%, respectively. Both models exhibited strong precision (95.31% for Random Forest and 95.71% for SVM), indicating minimal false positives. Logistic Regression also performed well, with an F1-score of 90.95%, but its slightly lower precision (88.79%) suggests a higher rate of false positives. Naive Bayes Classifier lagged behind with an F1-score of 88.15%, likely due to its assumption of feature independence, which may not hold for text data. Overall, the results highlight the superiority of transformer-based models like Bangla BERT for complex text classification tasks, while traditional models like Random Forest and SVM remain strong alternatives for resource-constrained scenarios.

Table III
PERFORMANCE OF EACH TRADITIONAL MACHINE LEARNING MODEL

| Model Name | Accuracy | Precision | Recall | F1-Score |
|---|---|---|---|---|
| Logistic Regression | 0.9151 | 0.8879 | 0.9321 | 0.9095 |
| Random Forest Classifier | 0.9420 | 0.9531 | 0.9186 | 0.9355 |
| Naive Bayes Classifiers | 0.8820 | 0.8808 | 0.8827 | 0.8815 |
| Support Vector Machine | 0.9400 | 0.9571 | 0.9095 | 0.9327 |
| Bangla BERT | 0.9565 | 0.9565 | 0.9524 | 0.9544 |

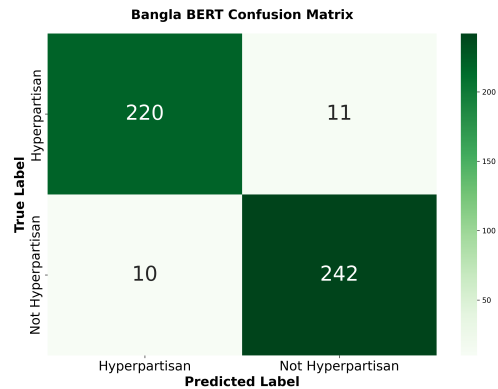

Figure 2. Bangla BERT Confusion Matrix

The confusion matrix analysis reveals particularly robust classification for positive instances, with 220 true positives and 11 false negative. For negative instances, the model shows good performance with 242 true negatives and 20 false positives. The overall F1-score of 0.9544 indicates balanced performance between precision and recall, suggesting that the model maintains consistency across both classes without significant bias.

## B. Explainability Analysis

The aim of this section is to explore the reasons behind the predictions made by our best model, BanglaBERT, which classifies news articles as either hyperpartisan or not hyperpartisan. By using explainable AI (XAI) techniques, we can gain insights into how the model reaches its conclusions, thereby building trust in its decisions. To achieve this, we employed LIME (Local Interpretable Model-Agnostic Explanation) to better understand the factors influencing BanglaBERT's predictions.

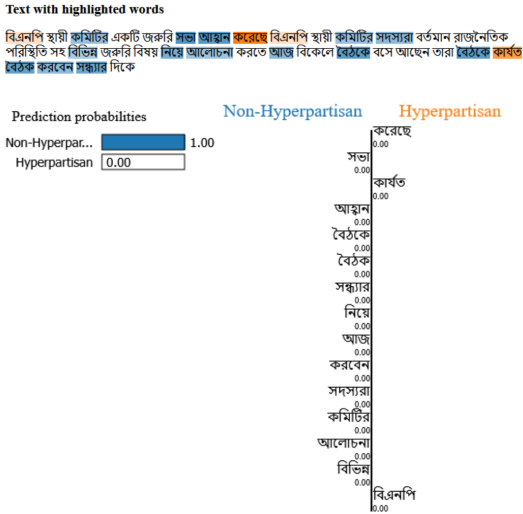

Figure 3. LIME explanation for a Non-Hyperpartisan news text in Bangla

Figures 3 and 4 illustrate LIME-based explanations of the Bangla hyperpartisan news classifier's decisions. In Figure 3, the model predicts the article as non-hyperpartisan with full confidence (1.00), highlighting neutral and procedural terms such as 'সভা' (meeting), 'আহ্বান' (call), and 'আলোচনা' (discussion). These words lack emotional or partisan language, signalling a balanced and factual tone. The visualisation demonstrates how the model relies on such neutral cues to distinguish non-hyperpartisan content.

Figure 4 shows a clearly hyperpartisan article, which the model labels with high confidence. Words like 'বিপরীত' (opposite), 'শাসন' (rule), and 'গণসংহতি' (mass solidarity) strongly influence the decision, reflecting emotional and political language. Some words like 'আকাঙ্ক্ষার' (aspiration) and 'সমন্বয়ক' (coordinator) slightly push against the hyperpartisan label, adding a more hopeful or neutral tone. This analysis shows that the model can pick up both strong and subtle clues in Bangla political texts.

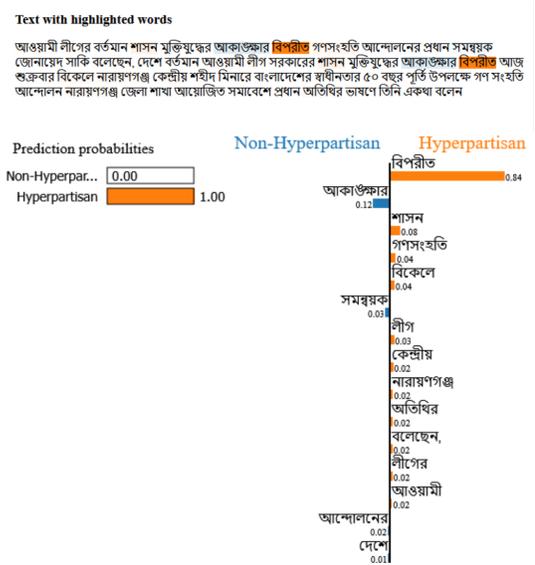

Figure 4. LIME explanation for a Hyperpartisan news text in Bangla

## C. Benchmark Comparison (Cross-lingual Contextualization)

To the best of our knowledge, this study represents the first attempt at hyperpartisan news detection in the Bangla language. Given the absence of prior work and the lack of publicly available benchmarks for this task in Bangla, there are no directly comparable models within the same linguistic context.

To contextualize our results, we reference several top-performing systems from the SemEval-2019 Task 4 on hyperpartisan news detection, which primarily focused on English and German corpora. Table 4 provides a comparative overview of F1-scores reported by these systems. While the top systems at SemEval-2019 achieved F1-scores ranging from 82.2% to 88.0%, our semi-supervised Bangla BERT-based model attained an F1-score of 95.44%.

Table IV
Cross-lingual Benchmark Comparison of Hyperpartisan News Detection Models

| System | Language | Approach | F1 Score (%) |
|---|---|---|---|
| Kiesel et al., 2019 (BERT) | English | Supervised BERT | 82.2 |
| Sanchez et al. (2019) | German | CNN + Attention | 88.0 |
| Ours (Bangla BERT) | Bangla | Semi-Supervised + LIME | 95.44 |

## VI. Conclusion & Future Work

This study is the first to address hyperpartisan news detection in the Bangla language using a semi-supervised fine-tuning approach with a Bangla-specific BERT model. Additionally, we are the first to apply LIME explainability in this context, offering interpretable and trustworthy predictions for Bangla text classification. These contributions establish a strong initial benchmark for future research on misinformation detection in low-resource languages.

By fine-tuning Bangla BERT and comparing it with traditional machine learning models, we demonstrated that transformer-based architectures significantly outperform Logistic Regression (F1: 90.95%), Random Forest (F1: 93.55%), and SVM (F1: 93.27%), achieving state-of-the-art performance (F1: 95.44%). Our study uses only political articles from BNAD, which may limit generalizability to other domains (e.g., health or economy) due to differences in vocabulary and style. Future research could enhance hyperpartisan detection by incorporating multimodal content such as audio and video. This may involve combining textual, acoustic, and visual signals using advanced multimodal architectures for improved accuracy.

The proposed model can support media watchdogs in identifying hyperpartisan content, integrate with real-time monitoring systems via APIs, and assist automated fact-checking pipelines by flagging potentially biased articles.